\documentclass{article}

\usepackage{microtype}
\usepackage{graphicx}
\usepackage{svg}
\usepackage{subfigure}
\usepackage{booktabs} 

\usepackage{hyperref}

\usepackage[arxiv]{icml2024}

\usepackage{amsmath}
\usepackage{amssymb}
\usepackage{mathtools}
\usepackage{amsthm}
\usepackage{mathrsfs}
\usepackage{enumitem}

\usepackage[capitalize,noabbrev]{cleveref}

\usepackage[textsize=tiny]{todonotes}
\usepackage{pifont}

\icmltitlerunning{Augmentations vs Algorithms}

\usepackage{diagbox}
\usepackage{custom}
\begin{document}

\twocolumn[
\icmltitle{Augmentations vs Algorithms: What Works\\in Self-Supervised Learning}



\icmlsetsymbol{equal}{*}
\icmlsetsymbol{intern}{+}

\begin{icmlauthorlist}
\icmlauthor{Warren Morningstar}{1}
\icmlauthor{Alex Bijamov}{1}
\icmlauthor{Chris Duvarney}{1}
\icmlauthor{Luke Friedman}{1}
\icmlauthor{Neha Kalibhat}{intern,2}
\icmlauthor{Luyang Liu}{1}
\icmlauthor{Philip Mansfield}{1}
\icmlauthor{Renan Rojas-Gomez}{intern,3}
\icmlauthor{Karan Singhal}{1}
\icmlauthor{Bradley Green}{1}
\icmlauthor{Sushant Prakash}{1}
\end{icmlauthorlist}

\icmlaffiliation{1}{Google Research}
\icmlaffiliation{2}{University of Maryland, College Park}
\icmlaffiliation{3}{University of Illinois at Urbana-Champaign}

\icmlcorrespondingauthor{Warren Morningstar}{wmorning@google.com}

\icmlkeywords{Machine Learning, ICML}

\vskip 0.3in
]



\printAffiliationsAndNotice{\googleIntern}  

\begin{abstract}
  We study the relative effects of data augmentations, pretraining algorithms, and model architectures in Self-Supervised Learning (SSL).  While the recent literature in this space leaves the impression that the pretraining algorithm is of critical importance to performance, understanding its effect is complicated by the difficulty in making objective and direct comparisons between methods.  We propose a new framework which unifies many seemingly disparate SSL methods into a single shared template.  Using this framework, we identify aspects in which methods differ and observe that in addition to changing the pretraining algorithm, many works also use new data augmentations or more powerful model architectures.  We compare several popular SSL methods using our framework and find that many algorithmic additions, such as prediction networks or new losses, have a minor impact on downstream task performance (often less than $1\%$), while enhanced augmentation techniques offer more significant performance improvements ($2-4\%$).  Our findings challenge the premise that SSL is being driven primarily by algorithmic improvements, and suggest instead a bitter lesson for SSL: that augmentation diversity and data / model scale are more critical contributors to recent advances in self-supervised learning.

\end{abstract}


\begin{figure*}[t]
    \centering
    \includegraphics[width=\hsize]{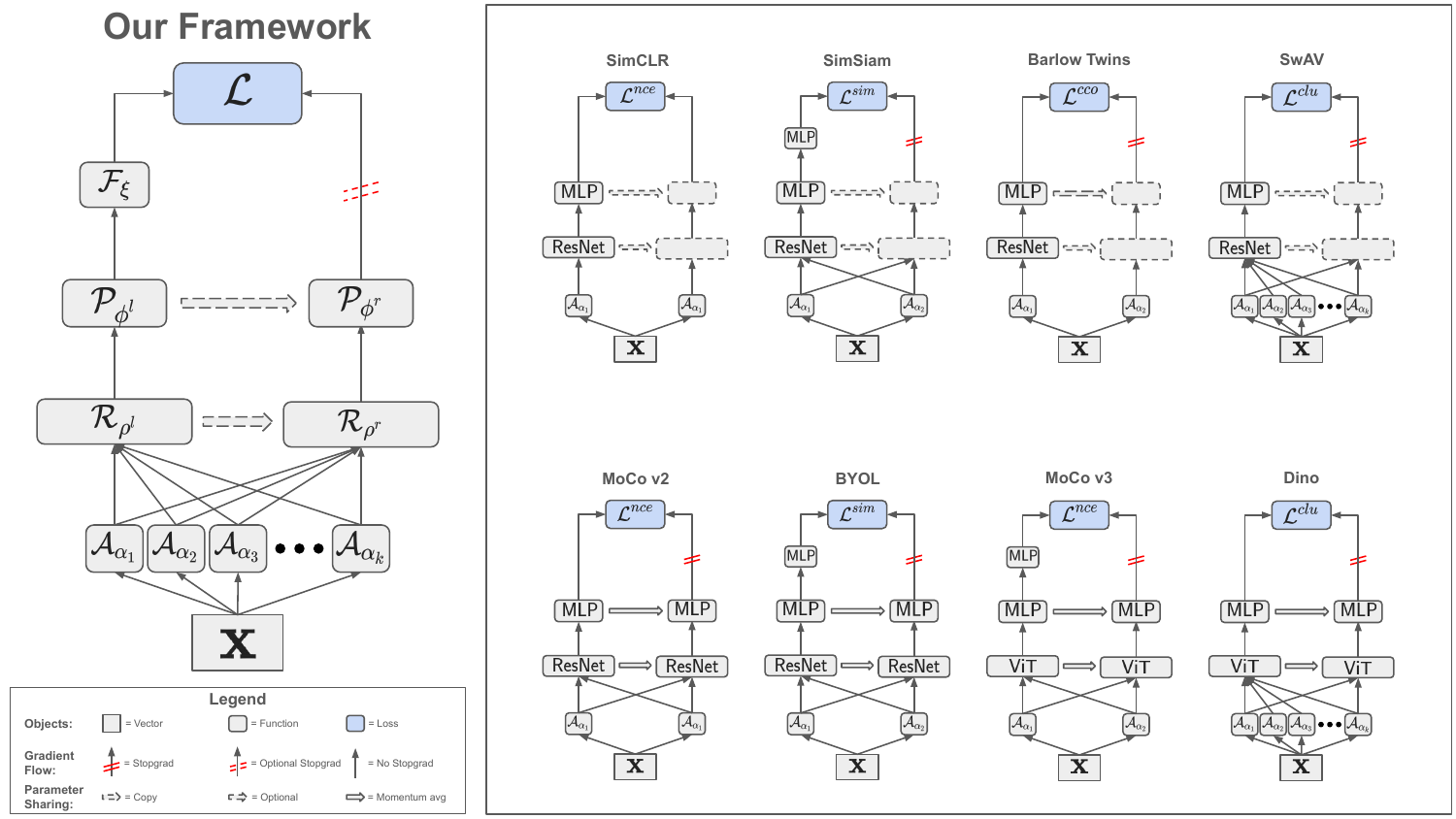}
    \caption{Our proposed SSL framework.  The right panel shows 8 popular SSL algorithms, each of which has seemingly different model graphs, network architectures, losses, and augmentation strategies.  We show that all of these methods are instances of a single unifed framework (left), each having different architecture hyperparameters for each module, augmentation hyperparameters, or losses.}
    \label{fig:ssl_algos}
\end{figure*}
\section{Introduction}\label{sec:introduction}
Data labelling is a difficult and expensive process \citep[e.g.][]{datalabeling1,datalabeling2,datalabeling3} that often is a primary factor limiting performance of machine learning (ML) models.  Self-supervised learning is an emerging ML paradigm that aims to reduce the need for such data labelling. In a self-supervised learning setup, models are ``pretrained" to perform an auxiliary task for which labels can be synthetically and efficiently generated \citep{simclr,byol,simsiam}. The aim of SSL is to devise pretraining tasks which teach the model to extract useful and general-purpose representations from input data.

Representations learned from self-supervised pretraining are often quite useful for downstream tasks.  This manifests itself in significantly increased label efficiency: models pretrained with SSL can often be adapted to a particular task with orders of magnitude fewer labels than models trained from scratch \citep{simclrv2}. Even more beneficial is the fact that the pretraining data need not be drawn from the downstream dataset; one can use a large unlabeled corpus of data to pretrain the model. This facilitates efficient training on small datasets, where standard supervised training typically causes large models to extract features which generalize poorly \citep{shortcut_learning}. Because of these advantages, a large and highly active area of research focuses on how to train self-supervised learning models that extract more informative, and thus more useful, representations.

To date, dozens of algorithms have been proposed, each of which builds upon previous techniques and improves the measured performance on many evaluation benchmarks.  However, direct comparisons between methods are mostly lacking, either because of the time and computational expense involved in making extensive comparisons or because of the difficulty in porting methods between different ML Frameworks \citep[e.g. PyTorch, TensorFlow, and JAX;][]{pytorch, tensorflow, jax}. As a result, it can be difficult to ascertain which techniques actually perform the most reliably, and why.

In this paper, we make the observation that many of the techniques which achieve a significant improvement in benchmark performance propose new advances in their data augmentation pipeline, or use larger model architectures (e.g. vision transformers, deeper projectors).  We dig deeper by running experiments on a variety of the existing pretraining methods, and come to the somewhat surprising conclusion that the largest driver in SSL model improvement has been increasing diversity in data augmentations.  Our main contributions are:

\begin{enumerate}[topsep=0pt,itemsep=-0.5ex,partopsep=1ex,parsep=1ex]
    \item We propose a new unified framework that generalizes many seemingly different SSL methods into a coherent template.
    \item We compare and experimentally replicate several popular SSL methods using our framework and identify that many recent advances in SSL involved enhancements in augmentations and architectures, in addition to commonly-emphasized advances in algorithms.
    \item We compare the relative performance impact of algorithms and augmentations, and observe the surprising result that augmentations constitute approximately $5\%$ of the reported performance improvement between early methods like SimCLR and modern methods like DINO, while model size appears to contribute an additional $2.3\%$.  Algorithmic improvements constitute a meager $2.1\%$, most of which can be attributed to the momentum encoder ($1.8\%$) and predictor ($0.3\%$).
    \item We demonstrate that, for both ResNet-based and ViT-based methods, the perceived performance gap between SSL algorithms can be eliminated through careful tuning of augmentations, momentum encoders, and prediction networks.
    \item Our findings indicate that SSL is not being driven primarily by algorithmic improvements, and suggest a bitter lesson for SSL: augmentation diversity and data / model scale are the crucial ingredients for advances.
\end{enumerate}

\section{Problem Setup}
\label{sec:problem}
In this section, we attempt to describe a general SSL problem formulation in order to ground our later discussion of the different methods in this space.  We start by introducing terminology.  SSL is a type of supervised learning where models generate their own training signals using the input data rather than using external labels.  This is related in many ways to unsupervised learning, as both are attempting to ascertain patterns in the data without using explicit labels. The main distinction between the two is that SSL uses some form of weak labels generated from the input data to induce learning, while unsupervised learning is generally focused on finding patterns within the data (using e.g. clustering or dimensionality reduction).

The goal of self-supervised pretraining is to learn an encoder; an ML model which leverages a large unlabeled dataset to extract common features from the input examples.  The encoder composes these features into a \textit{representation}; a lower-dimensional set of features which retains important information from the input while excluding superfluous information. The representations can then be used to perform a \textit{downstream task}, such as classification, object detection, semantic segmentation, or question answering for which some labels may exist but it may be difficult to generate labels at scale.

For most SSL algorithms, the encoder is pretrained to extract representations using some form of \textit{pretext task}; a training objective for which labels can be generated on the fly using the model and the input data. While in some cases (e.g. LLMs) this is sufficient, in many cases the task on its own does not cause the model to learn useful representations. In these cases, more diverse task labels are generated using some form of data augmentation, thus accomplishing the task requires the model either to undo it, measure it, or just ignore it.  While the pretext task trains the model to extract information from the data, the augmentations also help to inform the model as to which information is important and which is superfluous. 

Here, we attempt to mathematically describe a general SSL pretraining setup, albeit one which focuses on joint embedding methods; a subset of what has been studied in the literature.

\textbf{Notation and Assumptions:}
To disambiguate between the many different parameters, indices, functions and tensors used in the SSL methods under discussion, we establish the following conventions. We use caligraphic upper case letters (e.g. $\mathcal{R}$, $\mathcal{T}$) to denote parametric functions, Greek letters (e.g. $\rho$, $\theta$) to denote the parameters of those functions, and Greek subscripts to indicate dependency of the function on particular parameters (e.g. $\mathcal{R}_{\rho}$, $\mathcal{T}_{\theta}$).  We use lower case letters (e.g. $x$, $z$) to denote row vectors and their components, bold lower case letters (e.g. $\xmat$, $\zmat$) to denote rank-2 tensors, and bold upper case letters (e.g. $\xtens$, $\ztens$) to denote rank-3 tensors.  We use lower case superscripts as identifiers, lower case subscripts to denote an index in a tensor, and upper case superscripts to denote the number of elements over an index.  We try to reduce ambiguity when indexing by using $m$ and $n$ as indices in the dataset or minibatch, $i$ and $j$ as vector indices, and $k$ and $l$ as indices in a ``views" dimension used in joint embedding SSL.  We also reserve the indices $o$ and $p$ as indices to use as needed.  

We start with a dataset $\textbf{d}$, which is composed of unlabeled example vectors $x$, i.e., $\textbf{d} = \{x_n\}_{n=1}^{N}$.  Our goal is to train a model to generate representations $r = \mathcal{R}_{\rho}(x)$, where $\mathcal{R}_{\rho}: \mathbb{R}^{X}\rightarrow \mathbb{R}^{R}$ is our encoder, a function with parameters $\rho$ that converts each input datum into its corresponding representation with dimensionality $R$.  We further define an augmentation $\xp=\mathcal{A}_{\alpha}(x, \mathscr{R})$, where $\alpha$ represents a set of augmentation model hyperparameters (the types, strengths, and probabilities of augmentations being applied to x), and $\mathscr{R}$ is a source of random stochasticity which allows our augmentation model $\mathcal{A}$ to generate a different augmentation for each input $x$ passed to $\mathcal{A}$. Similar to most joint embedding SSL works described above, we also use a projector ($\mathcal{P}:\mathbb{R}^{R}\rightarrow \mathbb{R}^{I}$) to compute a lower-dimensional embedding from our representation $z = \mathcal{P}_{\phi}(r)$, where $I$ is the dimensionality of the embedding, and we use a predictor with parameters $\xi$ to generate a prediction $\hat{z}$ of the embedding, (i.e. $\hat{z}=\mathcal{F}_{\xi}(z)$).  For ease of notation, we further chain the projector and encoder together to construct a Tower $\mathcal{T}_{\theta}: \mathbb{R}^{X}\rightarrow\mathbb{R}^{I}$ where $\theta=\{\rho, \phi\}$ such that: \begin{equation}z=\mathcal{T}_{\theta}(x)=\mathcal{P}_{\phi}(\mathcal{R}_{\rho}(x))\end{equation}

To construct a forward pass, we subsample our dataset to get a minibatch of size $M$: $\xmat=\{x_m\}_{m=1}^M$. We then use our augmentations to generate $K$ different views of each example in the minibatch, \begin{equation}\xptens = \mathcal{A}_{\alphamat}(\xmat, \mathscr{R}) = \{\mathcal{A}_{\alpha_k}(\xmat, \mathscr{R})\}_{k=1}^K,\end{equation} where we assume each set of augmentation parameters $\alpha_k$ may be different (i.e. different views of a single input can have different distributions of augmentation hyperparameters), but that each set of parameters is shared between all elements of $\xmat$ (and thus the $k$-th view of all examples are generated using the same augmentation hyperparameters).  Note that even though the hyperparameters are the same for individual views, the exact chain of augmentations applied to each input is different due to the source of stochasticity $\mathscr{R}$.  We then compute the projection of each view from two towers \begin{equation}\ztens^{l}=\mathcal{T}_{\theta^{l}}({\xptens}),\end{equation} and \begin{equation}\ztens^{r}=\mathcal{T}_{\theta^{r}}({\xptens}),\end{equation} where we use the superscripts $l$ and $r$ to indicate the ``left" and ``right" towers respectively for the sake of intuitively distinguishing between them.  Note that the parameters $\theta^l$ and $\theta^r$ need not have the same dimensionality in our framework, but most prior work does assume the same dimensionality, and we thus have restricted our experimental evaluation to this case.  Finally, we pass the left tower outputs to the predictor to get a prediction \begin{equation}\predstens=F_{\xi}(\ztens_{l}).\end{equation}  Note that for some methods, the prediction network is just an identity function, and in many cases there is a stop gradient applied to $\ztens^{r}$ so gradients do not pass backwards through the right tower.

Once we have our tower outputs $\ztens^{r}$ and $\predstens$, we treat our right tower projections as labels and compute the loss of our left tower predictions $\mathscr{L}(\predstens, \ztens^r)$.  Note that this loss is itself the sum of several losses computed between paired views of the minibatch, i.e., \begin{equation}\mathscr{L}(\predstens, \ztens^r) = \frac{\sum_{k, l \text{=}1}^{K} \mathcal{M}(k, l)\mathcal{L}(\predsmat_{k}, \zmat_{l}^{r})}{\sum_{k, l\text{=}1}^{K}\mathcal{M}(k, l)},\end{equation} where $\mathcal{M}: \{1,\ldots,K\}^{2}\rightarrow\mathbb{R}$ weights the contribution between different pairs of views to the total loss.  The function $\mathcal{L}: \mathbb{R}^{M\times2\times I}\rightarrow\mathbb{R}$ is a nonlinear loss function defined between the predictions from a particular view of the batch, and the right tower projections of another view of the batch, and is used to specify the objective of the pretext task. We optimize the parameters of our model to minimize $\mathscr{L}$ using some variant of minibatch Stochastic Gradient Descent.  For some methods, the right tower is not trained by SGD and is instead treated as a momentum encoder \citep{moco}, in which case we perform a momentum update to the momentum encoder parameters \begin{equation}\theta^r = \epsilon \theta^{l} + (1-\epsilon)\theta^r\end{equation} at the end of each training step (note that this is only possible when the parameters have the same dimensionality).  This process repeats for several hundred epochs, after which we take one of the encoders and use it to generate representations for downstream tasks.

This framework is shown in \cref{fig:ssl_algos}. We identify 3 categories of ways in which methods differ.  The first is architectures; changes in the specific neural networks used in $\mathcal{T}_{\theta}$ or $\mathcal{F}_{\xi}$ such as wider ResNets, ViTs or deeper projection networks.  The second is augmentations; changes to $\mathcal{A}_{\alphamat}$ such as additional views of the input.  The third is algorithms, which we use to describe other changes to the SSL computation graph such as the use of momentum encoders and prediction networks, or new losses $\mathcal{L}$, (i.e. new pretext tasks).

Over the last several years, most of the attention in the SSL literature has focused on the algorithm, with a particular emphasis on $\mathcal{L}$ \citep[e.g.][]{byol, barlowtwins, swav, dino, msn, vicreg}. Many works also show that larger architectures improve performance \citep[e.g.]{simclr, byol,mocov3,swav,dino,msn}. Comparatively little explicit attention has been paid to the effect of new augmentations, with only \citet{swav, dino} showing the performance impact of their proposed augmentation, and generally concluding that it only really improves performance in combination with certain losses. However, side-by-side comparisons are difficult to implement and expensive to run, leading most empirical comparisons to be made against previously reported numbers on standard benchmarks. Since, new methods tend to adopt new architectures or propose new augmentations in conjunction with inventing new algorithms, it is generally unclear which aspects of each method (augmentations, algorithms, or architectures) are driving improvement.  Our goal is to study all of these methods jointly to answer this important question and to better infer the relative importance of different elements in self-supervised pretraining.
\begin{table*}[tb]
    \centering
    \begin{tabular}{l|cccccc}
        \toprule
        \diagbox[width=10em]{Augmentations}{Method} & SimCLR & BYOL & MoCo v2 & SwAV & DINO & MoCo v3  \\
        \midrule
        Crop & $43.6 \pm 0.3$ & $65.8 \pm 0.2$ & $63.2 \pm 0.1$ & $21.4 \pm 0.1$ & $49.8 \pm 0.7$ & $65.1 \pm 0.6$ \\
        Crop + Color & $65.4 \pm 0.2$ & $69.6 \pm 0.5$ & $69.2 \pm 0.3$ & $67.2 \pm 0.1$ & $58.5 \pm 0.9$ & $68.1 \pm 4.2$ \\
        SimCLR & $68.7 \pm 0.2$ & $71.5 \pm 0.3$ & $70.8 \pm 0.3$ & $71.0 \pm 0.3$ & $62.4 \pm 0.5$ & $74.2 \pm 3.0$\\
        BYOL & $69.5 \pm 0.4$ & $74.2 \pm 0.5$ &  $72.5 \pm 0.7$ & $70.7 \pm 0.2$ & $68.0 \pm 0.4$ & $77.4 \pm 0.2$\\
        Multi-crop &  $71.2 \pm 0.2$ & $74.3 \pm 0.9$ & $74.1 \pm 0.3$ & $74.6 \pm 0.3$ & $78.0 \pm 0.2$ & $75.9 \pm 1.2$ \\
        \bottomrule
    \end{tabular}
    \caption{\textbf{Effect of augmentations:} We explore the impact of increasingly diverse augmentations on downstream task performance, and report the top-1 accuracy observed on ImageNet found using linear probing. We find that increasing augmentation diversity has a significant positive impact on performance in almost all cases, suggesting that all methods benefit from increasing augmentation diversity.}
    \label{tab:augmentations}
\end{table*}
\section{Related Work}\label{sec:background}

Broadly speaking, SSL harnesses the machinery of supervised learning to accomplish the goals of unsupervised learning by utilizing the input data to generate labels for the pretext task. One widespread way of creating nontrivial labels for SSL is by leveraging data augmentations. Early SSL methods try to predict the parameters governing the augmentation, or to undo its effects.  Some examples of this are image colorization \citep{cic}, where the model is trained to restore the color in an image, masked autoencoding \citep{orig_mae, mae}, where the model is trained to fill in missing content, and rotation prediction \citep{rotnet} where the model is given a rotated image and asked to predict its rotation.  In all cases, the model is given some modified input data and trained to learn trends in the data (such that grass is green, or that the ground is always at the bottom of an image) that may aid in accomplishing the pretext task.   Often it is the case that these trends also aid the model in accomplishing the downstream task, and thus SSL provides a way to train a large model without requiring access to large labeled datasets.

Recent SSL approaches rely on training a model to perform pretext tasks between multiple ``views" of a single input, following the general framework presented in \cref{sec:problem}.  Methods differ mainly in what they use for each of the components during pretraining. The dominant focus is on improving the algorithms, with most methods devising a new formulation for the pretext task loss $\mathcal{L}$ or adding one of the modules described in \cref{sec:problem}.  Contrastive Predictive Coding \citep[CPC;][]{cpc} was a pioneering work of the dual encoding or joint embedding methods, and was also the first to use the widespread InfoNCE loss as their training objective.  Dropping view indices $k$ and $l$, this loss can be expressed in terms of batch dimensions $m$ and $n$ as follows:
\begin{equation}\label{eq:infonce}\mathcal{L}^{nce}(\predsmat, \zmat^{r}) = -\frac{1}{M}\sum_{m=1}^M\log\Softmax\left(\frac{\preds_m {\zmat^r}^{\top}}{\tau}\right)_m \end{equation} where $\Softmax: \mathbb{R}^{P}\rightarrow\mathbb{P}^{P}$ is the Softmax function, which takes a vector of arbitrary dimensionality and maps it to the probability simplex: \begin{equation}\mathcal{S}(x)_{o} = \frac{\exp{x_{o}}}{\sum_{p=1}^P \exp{x_p}}\end{equation}
In CPC, the loss is applied between extracted tokens in the representation, which is different from the formulation in \autoref{eq:infonce}. Later works such as MoCo \citep{moco} used $\mathcal{L}^{nce}$ as written above, where the loss is computed between examples in a batch rather than between the tokens extracted for a single example.  MoCo was also the first work to use a momentum encoder as the right tower, which has since become widely adopted.  \citet{cpcv2} improved on CPC by adding augmentations to each patch in the input.  SimCLR \citep{simclr} improved on all of these by adding a projection network and computing the loss in the projected space.  BYOL \citep{byol} proposed to use $\mathcal{L}^{sim}$ as the pretraining objective, where \begin{equation}\label{eq:lsim}\mathcal{L}^{sim}(\predsmat, \zmat^{r}) = \frac{1}{M}\sum_{m=1}^M \|\preds_{m} - z_{m}^{r}\|_2^2\end{equation} removes the contrastive penalty used in $\mathcal{L}^{nce}$ and just maximizes the similarity.  In order to prevent representation space collapse they combined the momentum encoder from MoCo and the projector of SimCLR, and added a prediction network, and found that in addition to preventing collapse, the model performance could be improved relative to SimCLR.  Concurrently, SwAV \citep{swav} reinterpreted the embeddings as the logits of a clustering model defined by a set of shared ``prototypes" (parameterized using the last layer in the projector), and proposed to use $\mathcal{L}^{clu}$ as the pretraining objective, where 
\begin{multline}\label{eq:lclu}\mathcal{L}^{clu}(\predsmat, \zmat^{r}) = -\frac{1}{M}\sum_{m=1}^M\sum_{i=1}^{I}\Sinkhorn^n\left(\frac{\zmat^{r}}{\tau^r}\right)_{mi}  \\ \log\Softmax\left(\frac{\preds_m}{\tau^l}\right)_{i}\end{multline}
 In this formulation, $\tau^r$ and $\tau^l$ are temperature hyperparameters used to define the hardness of the clustering.  The function $\Sinkhorn:\mathbb{R}^{M\times I}\rightarrow\mathbb{R}^{M\times I}$ is a recursive function with finite iterations which performs the Sinkhorn-Knopp algorithm and outputs an (approximately) doubly stochastic matrix \citep{sinkhorn}: \begin{equation}\Sinkhorn(\xmat)^n = \begin{cases}\Sinkhorn^{n-1}\left(\mathcal{V}(\mathcal{U}(\xmat))\right) & \text{if } n > 0 \\ \exp\left(\xmat\right) & \text{otherwise}\end{cases}\end{equation} where the functions $\mathcal{U}: \mathbb{R}^{M\times I}\rightarrow\mathbb{R}^{M\times I}$ and $\mathcal{V}: \mathbb{R}^{M\times I}\rightarrow\mathbb{R}^{M\times I}$ normalize the input matrix over rows and columns respectively. \begin{equation}\mathcal{U}(\xmat)_{mj} = \frac{x_{mj}}{\sum_{n}x_{nj}}\end{equation} \begin{equation}\mathcal{V}(\xmat)_{mj} = \frac{x_{mj}}{\sum_{i}x_{mi}}\end{equation} The Sinkhorn normalization is used to interpret its input as clustering assignment probabilities (since the columns sum to 1) while also trying to equipartition the inputs between clusters (since the rows approximately sum to 1). \citet{swav} showed that using $\mathcal{L}^{clu}$ improved the model performance relative to SimCLR.  \citet{barlowtwins} found that the performance of SimCLR could also be exceeded without the need for explicit contrasting with negative samples by computing a loss over the cross correlation matrix: \begin{multline}\mathcal{L}^{cco}(\predsmat, \zmat^{r}) = \sum_{i=1}^I\sum_{j=1}^I \delta_{ij} (1 - \mathcal{C}(\predsmat, \zmat^{r})_{ij})^2 \\ + \lambda(1-\delta_{ij})\mathcal{C}(\predsmat, \zmat^{r})_{ij}^2\end{multline} where $\delta_{ij}$ is the Kroeneker delta.  The function $\mathcal{C}$ computes the cross correlation matrix of the minibatch
\begin{equation}
    \mathcal{C}(\predsmat, \zmat)_{ij} = \frac{\sum_{m}\preds_{mi}z_{mj}}{\sqrt{\sum_{m}(\preds_{mi})^2}\sqrt{\sum_{m}(z_{mj})^2}}
\end{equation}
The effect of $\mathcal{L}^{cco}$ is to make the embeddings of positive pairs similar while minimizing the redundancy between each component in the embeddings.  \citet{vicreg} showed that models trained using $\mathcal{L}^{sim}$ could avoid collapse without the tricks used in BYOL and SimSiam \citep{simsiam} by instead adding regularization penalties on the variance and covariance of the embeddings and predicted embeddings. There are also multiple works which show that assimilating the modules proposed in other work can improve the performance of many of these methods \citep{mocov2, dino, mocov3, simclrv2, importanceofasymmetry}. 

As algorithms have improved, so has the downstream task performance, as evidenced by the observed performance on standard benchmarks such as ImageNet linear classification (48.7\% with CPC to 81\% with MoCo v3). However, as algorithms have improved, so too have architectures and augmentations. On the architectures front, recent works have started using larger models such as wide ResNets \citep{wideresnet} or large vision transformers \citep{vit} as the encoder, along with deeper projection networks \citep[e.g.][]{mocov3, dino, simclrv2}.  Other works \citep{simclr, byol, swav, randfield, sassl, fda} have increased the diversity of augmentations used to create the views being passed to the towers.  Even so, much of the discussion surrounding these advances has focused on aspects of the algorithm \citep[see e.g.][]{cookbook}, with comparatively little attention being paid to augmentations.

We make a few observations from studying this prior work using our framework. First, we notice that while no two of the methods described above are identical, it is evident that there are many similarities between them.  Most methods differ in the specifics of their algorithm, and may also differ slightly in terms of their architecture or augmentations. Second, in light of the first observation we find that the reported empirical comparisons against prior methods in the literature are not always straightforward to interpret. Many empirical comparisons are intended to distinguish between algorithms but simultaneously vary architectures or augmentations (or both). Finally, we observe that there is a lot that has not yet been studied, even in the space of currently existing model components.  Our framework not only makes it possible to identify precisely how two methods differ from each other, but it further allows us to make fair comparisons between them while simultaneously identifying which components of each method produce a meaningful impact on performance.  Our aim in this paper is to study a variety of widely used methods in order to identify trends that will provide clarity as to the factors underlying advancement in the field.  Future work will hopefully extend this view to include other self-supervised paradigms such as generative models \citep[e.g.][]{mae} or joint embedding predictive architectures \citep[e.g.][]{ijepa}.

\begin{table*}[t]
    \centering
    \begin{tabular}{cc|cccccc}
        \toprule
        \multicolumn{2}{c|}{Algorithm}& \multicolumn{6}{c}{Method}  \\
        \midrule\midrule
        Predictor & Momentum & SimCLR & BYOL & MoCo v2 & SwAV & Dino & MoCo v3\\\midrule
        \xmark & \xmark & $69.5 \pm 0.4$ & 0.1 & $69.7 \pm 0.2$ & $74.6 \pm 0.3$ & $75.8\pm0.2$ & $75.5\pm0.3$ \\
        \xmark & \checkmark & $71.6 \pm 0.1$ & $0.1$ & $71.5 \pm 0.2$ & $60.9 \pm 1.4$ & $78.0 \pm 0.2$ & $76.8\pm 0.2$\\
        \checkmark & \xmark & $69.6 \pm 0.2$ & $17\pm 9$ & $70.9\pm0.1$ & $74.4 \pm 0.2$ & $0.2\pm0.1$ & $75.7\pm0.1$ \\
        \checkmark & \checkmark & $72.0\pm 0.2$ & $74.2 \pm 0.4$ & $72.5 \pm 0.6$ & $57.1 \pm 1.2$ & $77.6 \pm 0.1$ & $77.4 \pm 0.2$ \\
        \bottomrule
    \end{tabular}
    \caption{\textbf{Algorithm ablations:} We explore the effect of prediction networks and momentum encoders on performance to diagnose how much of the improvement between methods is due to these enhancements.  We find that momentum encoders and prediction networks do not uniformly cause improvement, with clustering methods observing a  performance degradation from adding prediction networks.  Contrastive methods universally see an improvement from both. Similarity based methods appear to require a prediction network, with the models not using one observing representation space collapse.}
     \label{tab:prediction_network}
\end{table*}
\section{Experiments}\label{sec:experiments}
In Section~\ref{sec:problem} we described a framework that unifies many existing SSL techniques.  In this section, we use this framework to study the relative impact of augmentations and algorithms on the performance on downstream tasks.

\textbf{Experimental Framework:} Our experimental framework follows the setup detailed in Section~\ref{sec:problem}.  While the main idea underlying the framework is ML platform independent, we implemented the framework and ran our experiments in TensorFlow \citep{tensorflow}.

{\noindent \bf Datasets:}  In order to ground these results on previous experimental work in this space, we constrain our focus to pretraining using the ImageNet Dataset \cite{imagenet}, as this is the standard pretraining dataset for most work on SSL.  To keep the comparisons simple and to focus on the quality of the representations learned via pretraining, we also evaluate on ImageNet.  Due to the number of comparisons being made between methods, we did not compare performance in other scenarios like transfer learning, since most prior work seems to find that the ImageNet downstream task performance is a reasonable proxy for the transfer learning performance.

{\noindent \bf Models and Algorithms:}  To keep the comparisons manageable, we constrain our scope to instance-based joint embedding methods. This means that methods which perform patch-based dual encoding (e.g. CPC, Ibot, I-JEPA) or include generative models (e.g. MC-SSL, MAE) were not studied in this paper.  We further narrow our scope to methods which reported significant advances over the previous state of the art and which also proposed the use of either new augmentations or new architectures.  This means that some methods (Barlow Twins, SimSiam, VICReg) were not considered since they did not appear to move the current state of the art at their time of publication.   In Section~\ref{sec:results} we show that the pretraining task has relatively little impact on downstream performance compared to augmentations, architectures, or other features of the algorithm. Thus it is plausible that these methods did not improve performance because their main contribution was a new task.

The methods we studied in our experiments were SimCLR, BYOL, SwAV, MoCo-v2, DINO, and MoCo-v3.  Between them, we study a mix of different architectures (ResNet, ViT backbones, different MLP architectures for the projector) and algorithms (Contrastive, Clustering, and Self-Distillation tasks, momentum encoders, and prediction networks).  Note that in the interest of consistency with previous works, unless otherwise specified, we use the same model configurations as were used in the original implementation of each method. The exact configurations for each method can be found in the supplement.

{\noindent \bf Pretraining Protocol:}  Unless otherwise specified, we follow the pretraining setup specified in each individual method, since this setup should already be optimized for performance. We note that this means that some methods have embellishments that other methods lack (e.g. weight decay or temperature schedules), and it is therefore possible that we could eke out a minor amount of additional performance for other methods.  Because tuning the configurations to this degree would have required a prohibitive amount of compute, we will defer this tuning for now with the expectation that this will not significantly change our conclusions, since the methods which are most performant are the ones which use these extra embellishments.  We also present a few minor ablations showing the impact of some of these embellishments on performance, and find that in most cases the effects are relatively minor.

{\noindent \bf Evaluation Protocol:}  Unless otherwise specified, we replicate the evaluation protocol used to evaluate each individual method, since those were optimally tuned for that particular learning algorithm and thus put the best foot forward for each.  We found that there was no single evaluation protocol that worked uniformly well for all methods, and that different learning rates, weight decays, optimizers and numbers of epochs were necessary to replicate the published performance of each method.  

{\noindent \bf Augmentations:} To study the effect of augmentations on the performance of each method, we consider 5 different augmentation strategies: random cropping, crop and color jitter, SimCLR augmentations \cite{simclr}, BYOL augmentations \cite{byol}, and multi-crop augmentations \cite{swav}. We note that this follows the historical precedent of pretraining augmentations, and represents a steady increase in the diversity of views of the training data being seen by models.  The specifics of each configuration can be found in the supplementary material.
\section{Results and Discussion}\label{sec:results}
To determine the degree to which performance varies for a single set of hyperparameters, we run five independent trials, each using different random seeds.  We report the mean and standard deviation for each setting.  We note that this proved important, since certain settings can exhibit significant variance in the measured outcome.  To the best of our knowledge, this has not been studied in previous work.

\textbf{The Effect of Augmentations:} \cref{tab:augmentations} shows the results of varying the augmentation diversity while keeping architectures and algorithms consistent with the implementations of each published method.  We find that increasing augmentation diversity has a significant impact on the downstream task performance across all settings, improving the measured accuracy by an average of 23\% relative to the least diverse strategy, and an average of 5\% relative to SimCLR augmentations.  BYOL augmentations produced a fairly strong improvement relative to SimCLR augmentations, averaging 2.3\% across settings.  This is surprising, given that the set of BYOL and SimCLR augmentations are mostly the same, with BYOL having only added solarization as a new augmentation, and only slightly changing the hyperparameters of the blurring, cropping, and color augmentation.  This suggests that subtle and often understated changes in the augmentation strategy can substantially affect results in the SSL literature. Notably, BYOL augmentations do not increase training time, in contrast to multi-crop, which can increase the training time significantly.  We also find that the impact of BYOL augmentations is more significant for models using a momentum encoder, with SimCLR and SwAV seeing a smaller benefit than other methods.  This suggests that the effect of some algorithmic changes may be offering more potential improvement via diverse augmentations. 

We also find a similar average improvement ($2.5\%$) when using multi-crop.  However, here the improvement does not appear to be as universal, with larger gains for DINO ($+9.6\%$) and SwAV ($+3.6\%$). We note that the observed improvement on DINO is significantly larger than the improvement reported in \citet{dino}, who observed only a 3.6\% improvement using ViT-S.  This could be caused by several factors, including differences between ViT-S and ViT-B, or (more likely) covariance between the augmentation configuration and the pretraining setup.  We speculate that it is the latter, and note that this would make DINO the only method in which we observed such significant covariance.  We find that BYOL does not receive a statistically significant improvement in performance when using multi-crop, though we also note that the best performing run for BYOL used multi-crop.  We do not observe the performance degradation for BYOL reported by \citet{dino}, likely because we followed the BYOL pretraining configuration and only varied augmentations, while they tried to make a more direct comparison by changing the architecture to ViT-S.  This stresses the need to make comparisons in a self-consistent framework, as differences between the implementations on different platforms can seemingly contribute significantly to the observed performance.  We also find that MoCo v3 is negatively impacted by using multi-crop, though this is mainly because several trials had training instability, similar to the instability originally reported in \citet{mocov3}. Identifying the root cause of this instability and correcting it could improve performance for this method.

\textbf{The Effect of Algorithms:} In addition to the impact of augmentations, we also notice several other trends from \cref{tab:augmentations}. We first notice that even with improved augmentations, a performance gap still exists between methods (4\% on average relative to DINO). We also notice that the performance gap is significantly smaller (0.6\%) for methods using the same encoder architecture, and that it is largest for SimCLR, which could potentially receive a performance benefit from a momentum encoder or prediction network.  However, it is not clear if other methods would also receive the same benefit.  We therefore ran an additional ablation for all methods, adding or removing the momentum encoder and the prediction network where appropriate, in order to measure the effects of each algorithmic enhancement individually.

The results of this experiment are presented in \cref{tab:prediction_network}.  We find that the prediction network introduces an average improvement of only 0.3\% and therefore appears to be non-essential.  The exceptions are BYOL, where the prediction network is critical, and clustering methods, where the prediction network is detrimental to performance.  For BYOL, \citet{byol} hypothesize that the prediction network helps to make the collapsed representation (an undesirable global minimum of the pretraining objective) an unstable equilibrium that is not reached.  It is not clear to us why prediction networks are universally detrimental to clustering methods.  The momentum encoder is correspondingly more important, with most methods receiving a small but not insignificant (1.8\% on average\footnote{These calculations used only a subset of the experiments, for a full discussion on the methodology see \cref{sec:computing_averages}}) performance improvement.  The exception is SwAV where we find that the momentum encoder is detrimental to performance, though the results from \citet{dino} suggest that it could probably be tuned to perform equivalently.  Importantly, we find that SimCLR and MoCo v2 perform equally well when both a prediction network and momentum encoder are used, suggesting that the discrepancy between SimCLR and SwAV is mainly due to the lack of a momentum encoder and predictor for SimCLR.

\textbf{The Effect of Architectures:} One fairly well established prior result in the literature is that increased model size and compute can significantly improve performance \citep[e.g.][]{dino, mocov3,dinov2,ijepa}. Replicating these results is outside of the scope of our study, but we are still able to approximately measure the impact of switching from ResNet-50 to ViT-B (a roughly threefold increase in the number of parameters) with fixed augmentations and algorithms using our experimental results.  We find an average improvement of 2.2\% from increasing the model size. When combined with the approximate improvements from augmentations (5\%), momentum encoders (1.8\%) and predictors (0.3\%) we are able to approximately explain the performance gap between DINO and SimCLR.

Given these results, it does not appear that the pretraining task contributes substantially to performance.  In particular, with proper tuning of augmentations, momentum encoders, and prediction networks, we find that all ResNet-based methods perform equivalently to within our margin of error.  We also find that MoCo v3 can achieve 78.2\% accuracy with proper tuning, exactly matching the reported performance of DINO.  This suggests that when all of the embellishments are stripped away, the pretraining task is unimportant and that performance gains are mostly driven by increasing scale (which causes an average performance improvement of 2.2\% when using fixed augmentations and algorithms), and increasing augmentation diversity (which causes an average improvement of 12.2\% across all augmentations, and a 5\% improvement from the SimCLR augmentation baseline). Augmentations cause the largest performance improvement in our experiments, suggesting that their importance to self-supervised learning may be significantly underestimated.

\section{Conclusions}\label{sec:conclusions}
We studied the relative impact of augmentations and algorithms on the quality of learned representations.  To do this, we proposed a framework which unifies dual encoder SSL methods, and enables us to identify specifically how each method differs from alternatives.  Using this framework we were able to independently vary the augmentations used during pretraining for several popular methods and showed that increasing augmentation diversity has been a significant driver of the advancement in SSL.  We also studied the impact of the algorithm, and found that while certain elements have strong effects on particular methods (e.g., momentum encoders and prediction networks), the general impact is smaller than that of augmentations.  We further find that the pretext task (training loss) has a minimal impact on downstream task performance.  This challenges the conventional wisdom that innovations in SSL algorithms have been the dominant drivers of downstream task performance and suggests instead that augmentations are much more important than previously thought.
\section{Impact Statement}
This work studies the relative effects of data augmentation and pretraining algorithms in SSL.  SSL is a generally useful technology with potentially significant societal impact. While our work studying and ablating data augmentations and algorithmic advances may not directly result in an immediate societal benefit, it does help improve understanding, explainability and trustworthiness of machine learning techniques. Furthermore, identifying and emphasizing the importance of data augmentations helps democratize the advancements of SSL by allowing wider audiences and applications to benefit not only from increased scale, which may not be feasible to implement in resource-constrained settings, but also from focusing on designing domain-appropriate and task-specific data augmentations.

\bibliography{references}
\bibliographystyle{icml2024}

\newpage
\appendix
\onecolumn
\section{Comparisons between models}
In this section, we aim to concretely identify how different methods are similar and how they are different.

\subsection{SimCLR}
SimCLR is the baseline we use for SSL techniques.  SimCLR uses a ResNet-50 for its encoder architecture, and a 2 layer MLP for the projector, where the MLP has a hidden dimensionality of 4096, an output dimensionality of 256, batchnorm after every dense layer, and ReLU activations.  It was trained for 1000 epochs using a batch size of 4096, the LARS optimizer \cite{lars}, and the InfoNCE loss.

\textbf{Relationship to MoCo v2:} SimCLR and MoCo v2 are similar in many details of their architecture and pretraining configuration.  The main differences between the two are that MoCo v2 uses a non-trainable momentum encoder for the right tower whose parameters are an exponential moving average of the left tower. SimCLR instead uses the same model for both towers.  SimCLR further allows gradients to propagate through both paths of the model (left and right paths).  The other major difference is that MoCo v2 originally used a queue of embeddings to keep the batch size small while still having a sufficiently large pool of negative samples.  However, if one can train with a large batch (e.g. 4096), there is not a significant benefit to using the queue of embeddings.  Other small details may differentiate them, however we found in our experiments that most of those produced no major impact on performance.

\textbf{Relationship to SwAV:} SimCLR and SwAV are also similar in many details of their architecture and pretraining configuration.  The major differences between them was that SwAV used $\mathcal{L}^{clu}$ rather than $\mathcal{L}^{nce}$, and that SwAV used multi-crop augmentations during pretraining while SimCLR did not.  However, there were many minor differences between them which ended up proving important in aggregate.  SwAV for example uses a different architecture for the projection head, having a generally lower dimensional feature space of 2048 and 128 for the two MLP layers respectively.  They also modified a significant number of elements of the pretraining and evaluation hyperparameters relative to SimCLR.

\textbf{Relationship to BYOL:}  Similar to MoCo v2, SimCLR and BYOL use the same architecture for their encoder and projector, and more or less use the same pretraining and evaluation setups.  The main differences between the two are that BYOL uses a momentum encoder (similar to MoCo v2), that BYOL adds a prediction network (using the same architecture is its projector) in order to prevent representation space collapse, and that BYOL trains using $\mathcal{L}^{sim}$.  The other subtle difference is that BYOL uses different augmentations during pretraining.  While the set of augmentations is largely the same, BYOL reduces the strength of brightness and contrast, and hue distortions in half, and reduces the saturation strength to a quarter of that used by SimCLR.  BYOL also adds solarization to the right view, reduces the probability of blurring of the left view to 0.1, and draws from a logarithmic distribution of axis ratios rather than a linear distribution.  While these differences are subtle, it actually had a significant impact on performance, causing a nearly 3\% improvement when the BYOL augmentations were used.

\subsection{BYOL}
\textbf{Relationship to MoCo v2:} MoCo v2 and BYOL use the same architecture for the encoder and projector, but BYOL adds a prediction network relative to MoCo v2. Otherwise the main difference between them was the pretext task, with BYOL minimizing $\mathcal{L}^{sim}$ and MoCo v2 minimizing $\mathcal{L}^{nce}$, and the changes to the augmentations used in pretraining BYOL.  All other differences between them did not appear to have a significant impact on performance.

\textbf{Relationship to SimSiam:} SimSiam tries to train models using the same loss as BYOL but without the momentum encoder.  This is accomplished by changing several elements of the architecture, namely introducing a dimensional bottleneck in the prediction network, and using a different projection network architecture (three 2048-d layers with batchnorm applied after all of them). SimSiam also appears to work better with smaller batch sizes, with large batch sizes observing a significant performance degradation.

\subsection{MoCo v2}
\textbf{Relationship to MoCo v3:} The main difference between MoCo v2 and v3 is the architecture and the pretraining hyperparameters.  MoCo v2 used a ResNet for the encoder and a 2 layer MLP for the projector, while MoCo v3 used a ViT and a 3 layer MLP for the projector.  MoCo v3 also added a predictor relative to the published implementation of MoCo v2, though our experiments seem to indicate that this had a relatively minor impact on performance.  Because MoCo v3 used a vision transformer instead of a ResNet, they also had to significantly adjust much of the pretraining hyperparameters, using AdamW \citep{adamw} as the optimizer, increasing the loss temperature parameter, lowering the right tower momentum, increasing the weight decay significantly, and lowering the learning rate significantly.
\subsection{SwAV}
\textbf{Relationship to DINO:}  Both SwAV and DINO train to minimize $\mathcal{L}^{clu}$, use multicrop, and do not use a prediction network.  The main differences between them is the architecture, where SwAV uses ResNet-50 and a 2 layer MLP for the encoder and projector respectively, where DINO uses a vision transformer for the encoder and a 3 layer MLP without batch normalization \citep{batchnorm} and with GELU activations for the projection head.  There are two algorithmic differences between them, namely that DINO uses a momentum encoder and that the loss for Dino uses a centering operator rather than the Sinkhorn-Knopp algorithm to equipartition the clusters.  However the latter of the two was found to have no significant impact on performance ($\pm0.1\%$) in their paper, well within the margin of random scatter that we observe.  Finally, DINO has significant tuning to their pretraining setup, employing temperature schedules to their loss and weight decay schedules for their optimizer.  We found in our experiments that the temperature schedule had a minor impact on performance ($-0.1\%$ for fixed temperature), but that the weight decay schedule had a significant impact, with performance degradations of $-3.5\%$ for $wd=0.04$, a performance degradation of $1.5\%$ for $wd=0.4$ and a performance degradation $1\%$ for $wd=0.1$, noting that their weight decay follows a cosine decay schedule from $0.04$ to $0.4$, and thus that the values we tried were the endpoints and the middle point of their decay range.
\subsection{DINO}
\textbf{Relationship to MoCo v3:} Dino and MoCo v3 have a mostly similar architecture, with the main architectural differences being that DINO uses stochastic depth in their encoder (though our experiments indicate that dropping stochastic depth has a neutral effect on perforamance), and has a lower dimensional projection head (2048 for the hidden dimensions rather than 4096) which does not use batch normalization.  We found that a higher dimensional projection head (4096) was detrimental to performance.  

The algorithmic differences between them are that they perform different tasks, with DINO minimizing $\mathcal{L}^{clu}$ and MoCo v3 minimizing $\mathcal{L}^{nce}$, and that Dino does not use a prediction network while MoCo v3 uses a 2 layer MLP with a configuration similar to that used in BYOL. One other important difference between them is the mechanism used to avoid training instability, which seems to plague transformer-based SSL methods.  DINO avoids instability by training using a small batch size (1024) and using gradient clipping, while MoCo v3 retains a large batch size but freezes the patch projection layer.  We find in our experiments that DINO receives a modest reduction in performance from removing gradient clipping ($0.5\%$), partially as a result of increased instability during training (as viewed by periodic spiking in the loss).  We also find that DINO collapses when a larger batch size is used (2048), suggesting that reducing the batch size was also important to reduce instability.  MoCo v3 tolerates a minor amount of instability, but we find that the instability grows when multi-crop is used, and that while lowering the batch size does reduce the training instability when multicrop is used, it ultimately results in a less performant model ($74.9\%$ from $75.9\%$). \section{Additional details of experiments}\label{sec:hyperparameters}

\subsection{Augmentation hyperparameters}

We used five separate augmentation hyperparameter configurations, set to progressively become more diverse.  They are listed below:
\begin{enumerate}
    \item \textbf{Crop:} This augmentation strategy returns 2 crops and provides no additional augmentation.
    \item \textbf{Crop + Color:} This augmentation strategy performs the previous cropping strategy and subsequently performs random color jitter with the strengths equal to that used in the SimCLR paper \citep{simclr}.
    \item \textbf{SimCLR:} This augmentation strategy follows the implementation in \citet{simclr} and sequentially performs random crop, color jitter, grayscaling, horizontal flipping, and blurring.
    \item \textbf{BYOL:} This augmentation strategy follows the implementation in \citet{byol}, which modifies some of the hyperparameters from the SimCLR strategy and adds random solarization to one of the 2 global views.
    \item \textbf{Multi-crop:} This augmentation strategy was defined in \citet{swav} and follows the implementation used in the DINO paper \citep{dino}.  It consists of BYOL augmentation for the global views, and an additional 10 "local views" generated using a hybrid of the SimCLR and BYOL augmentation strategies.
\end{enumerate} The exact hyperparameters used in each strategy can be found in \cref{tab:augmentation_hparams}.

\begin{table*}[tb]
    \centering
    \begin{tabular}{ll|lllll}
        \toprule
        \multicolumn{2}{c|}{Augmentation} & \multicolumn{5}{c}{Augmentation strategy} \\ \midrule
        Type & Parameters & Crop & Crop + Color & SimCLR & BYOL & Multi-crop \\ \midrule 
        Global Crop & area range & 0.08-1 & 0.08-1 & 0.08-1 & 0.08-1 & 0.25-1 \\
        & axis ratio range & 0.25-1.33 & 0.25-1.33 & 0.25-1.33 & 0.25-1.33 & 0.25-1.33\\
        & axis ratio distribution & uniform & uniform & uniform & logarithmic & logarithmic \\
        & output size & 224 & 224 & 224 & 224 & 224 \\
        & number of crops & 2 & 2 & 2 & 2 & 2 \\ \midrule
        Local Crop & area range & --- & --- & --- & --- & 0.08-0.25 \\
        & axis ratio range & --- & --- & --- & --- & 0.25-1.33 \\
        & axis ratio distribution & --- & --- & --- & --- & logarithmic \\
        & output size & --- & --- & --- & --- & 96 \\
        & number of crops & 0 & 0 & 0 & 0 & 10 \\ \midrule
        Color Jitter & Contrast & --- & 0.8 & 0.8 & 0.4 & 0.4 \\
        & Brightness & --- & 0.8 & 0.8 & 0.4 & 0.4\\
        & Saturation & --- & 0.8 & 0.8 & 0.2 & 0.2\\
        & Hue & --- & 0.2 & 0.2 & 0.1 & 0.1 \\
        & probability & 0 & 0.8 & 0.8 & 0.8 & 0.8\\ \midrule
        Grayscale & probability & 0 & 0 & 0.2 & 0.2 & 0.2 \\ \midrule
        Hflip & probability & 0 & 0 & 0.5 & 0.5 & 0.5 \\ \midrule
        Blur & kernel width & --- & --- & 0.1-2 & 0.1-2 & 0.1-2 \\
        & probability & 0 & 0 & 0.5 & [1.0, 0.1] & [1.0, 0.1, 0.5] \\ \midrule
        Solarize & threshold & --- & --- & --- & 0.5 & 0.5\\
        & probability & 0 & 0 & 0 & [0, 0.2] & [0, 0.2, 0]\\

        \bottomrule
    \end{tabular}
    \caption{\textbf{Augmentation Hyperparameters}: The parameter configs for augmentations used in our experiments. Dashed numbers indicate a range over which the numbers were sampled, and square brackets indicate different singular values used for each tower, where the first 2 values correspond to the global views and the final value is used for local views.}
    \label{tab:augmentation_hparams}
\end{table*}

\subsection{Model hyperparameters}
For each method, to the best of our ability, we replicate the setup used in the original paper. This means that SimCLR, MoCo v2, BYOL and SwAV used ResNet-50 as their encoder architecture, while DINO and MoCo v3 used ViT-B/16. All ResNet-50 methods did not deviate from the original published implementation in any hyperparameters.  For ViT methods, we used the standard ViT-B/16 configuration for both DINO and MoCo v3, meaning that we used 12 ViT blocks, 12 attention heads, 3072 hidden units in the MLP, a dropout probability of 0.0, an attention dropout probability of 0.0, and a layer norm stability constant of $10^{-6}$.  The patches were 16 pixels in both height and width and were projected to a 768 dimensional vector.  We used fixed sin-cos embeddings to encode positional information.  Both methods used a $cls$ token to extract the representation information during pretraining.  For DINO, we also used stochastic depth with probabilty 0.1 on the ViT, and for MoCo v3 we did not apply gradients to the patch projection layer \citep[following][]{mocov3}.

\subsubsection{Projector}
\textbf{SimCLR:}  SimCLR used a 2 layer projection head, with a hidden dimensionality of 4096 and an output dimensionality of 256.  SimCLR applied batch normalization after every dense layer, and then ReLU on all layers except for the last layer.  The activation was applied after batch normalization.

\textbf{BYOL:} BYOL largely uses the same projection head as SimCLR, where the main difference is that BYOL does not apply batch normalization after the final dense layer.

\textbf{MoCo v2:} For MoCo v2, we used the same projector architecture as BYOL to make more direct comparisons between the two and to make a better comparison between MoCo v2 and SimCLR. We note that this deviates from the original implementation of MoCo which uses a 2048 dimensional hidden layer, though the exact architecture of the MoCo projector is not entirely clear from reading the paper.  We find that the performance of MoCo v2 is largely equivalent to that reported in the paper, suggesting that any discrepancies have a minor impact on performance.

\textbf{SwAV:} For SwAV, we followed the implementation in \citet{swav} which uses a 2048 dimensional hidden layer, and a 128 dimensional embedding dimensionality, followed by a prototypes layer which l2-normalizes the input and then feeds it to a weight normalized dense layer with no bias.  SwAV uses batch normalization and a relu activation after the first hidden layer, but not after the dimensional bottleneck.

\textbf{DINO:} For DINO, we followed the implementation in \citet{dino}, which uses 2 2048 dimensional hidden layers, followed by a 256 dimensional bottleneck layer, followed by the prototypes layer.  However, unlike DINO, we found that we only needed to use 4096 prototypes to replicate the performance in \citet{dino}, most likely because we used sinkhorn normalization in the loss rather than centering.  The DINO projector did not use any normalization, and applied GELU activations \citep{gelu} after each dense layer.

\textbf{MoCo v3:} For MoCo v3, we followed the implementation in \citet{mocov3}, which uses two 4096 dimensional hidden layers, followed by a 256 dimensional output layer.  Following the published implementation, we used batch normalization after all layers in the MLP, including the final layer.

\subsubsection{Prediction network}
SimCLR, SwAV, and DINO do not use prediction networks.  MoCo v2 does not appear to use one, but we used one for MoCo v2 in our experiments in order to make more direct comparisons to BYOL and MoCo v3, both of which use prediction networks. In all cases, the prediction network uses the same architecture, which follows the BYOL projector. To be precise, the prediction network was a 2 layer MLP with 4096 hidden units and and output dimensionality of 256, using batch normalization and a relu activation after the hidden layer.

\textbf{Predictor Ablations:} When ablating the effect of the prediction network, we use the same predictor for SimCLR and SwAV as was used with other ResNet-50 methods. For DINO when using a predictor we removed the prototypes bottleneck from the projector and used a single prototypes layer for all outputs, which follows the implementation of the prototypes in \citet{msn}.  The predictor itself followed the template of the projection network (no batch normalization, gelu activations, 2048 hidden units), but following MoCo v3 we used a 2 layer prediction network rather than a deeper model.

\subsection{Optimization hyperparameters}
To the best of our abilities, we replicated the setups of each of the template methods.  The exception is DINO where we replicated the setup from \citet{wessl}, since they reported configurations that used Sinkhorn Normalization rather than the centering operation.  The exact configurations are as follows:

\textbf{SimCLR, BYOL, MoCo v2, and SwAV:}  All of these methods use the LARS optimizer \citep{lars}, a batch size of 4096, and a linear warmup + cosine decay learning rate schedule.  SimCLR, oCo v2 and SwAV used a peak learning rate of 4.8 (0.3 scaled by the batch size), while BYOL used a slightly lower learning rate (3.2, which is a base learning rate of 0.2 scaled by the batch size).  For BYOL, SimCLR, and MoCo v2, we used a weight decay of $1.5\times10^{-6}$, while for SwAV the weight decay was a bit lower ($10^{-6}$).  In all cases, we used a trust coefficient of 0.001, excluded biases and batch normalization parameters from weight decay and layer adaptation, and a traditional momentum of 0.9 (following the original implementations of all of these methods).  SwAV also used a slightly different learning rate schedule, where the learning rate was warmed up from 0.3, and cooled down to $4.8\times10^{-3}$.  All other methods started warmup from 0, and cooled down to 0.  All methods were trained for 1000 epochs, with the exception of SwAV which was trained for 800 epochs.  Note that this follows the implementations from each method.

\textbf{DINO:} We followed the DINO optimization configuration used in \citet{wessl}, which is mostly similar to the original setup from \citet{dino}, but slightly tuned to work without the centering operator.  They used a batch size of 1024, the Adam optimizer \citep{adam, adamw}, and a cosine decay schedule for both the learning rate and the weight decay, and with linear warmup for the learning rate.  For the learning rate, the warmup was initialized at 0.0625 of the peak learning rate, increased to a maximum value of 0.003 ($7.5\times10^{-4}$ scaled by the batch size) over 30 epochs, and decayed to $10^{-6}$.  We found that the end value did not matter significantly, so long as a sufficiently low value was used.  The weight decay schedule had no linear warmup, and increased from 0.04 to 0.4 following a cosine decay schedule.  We found that this had a significant impact on performance, which could not be matched using a fixed value for the weight decay.  We also did not apply weight decay to the biases or layer norm parameters, following the implementation in \citet{dino}.  Dino was trained for 400 epochs following \citet{dino}.

\textbf{MoCo v3:} We followed the optimization configuration used in \citet{mocov3}.  They used Adam as the optimizer with a constant weight decay of 0.1, and a learning rate which is warmed up from 0 to 0.0024 ($1.5\times10^{-4}$ scaled by the batch size) over 40 epochs, and then cooled back down to 0.

\subsection{Pretext task}
In all cases except DINO, we followed the original implementation of the pretext task, which we outline below.

\textbf{SimCLR and MoCo v2:} SimCLR and MoCo v2 used the same loss function ($\mathcal{L}^{nce}$ and the same temperature ($0.1$)

\textbf{BYOL:} BYOL used $\mathcal{L}^{sim}$ as the loss.  This loss has no hyperparameters to be tuned.

\textbf{SwAV:} SwAV used $\mathcal{L}^{clu}$ as the loss.  They used a temperature of 0.1 for the left tower and 0.05 for the right tower.  They performed 3 steps of Sinkhorn normalization to equipartition the examples between clusters.

\textbf{DINO:} DINO also used $\mathcal{L}^{clu}$ as the loss. They used a fixed temperature of 0.1 for the left tower, but used a temperature schedule for the right tower.  Following \citet{wessl}, we used a final temperature of 0.025 and an initial temperature of 0.05 with 3 steps of Sinkhorn normalization.

\textbf{MoCo v3:} Similar to MoCo v2, MoCo v3 used $\mathcal{L}^{nce}$ as the loss, but used a lower temperature (0.2 rather than 0.1).

\subsection{Downstream evaluation}
We used multiple different downstream evaluation configurations, as we found that deviations from a single template were necessary to replicate the published performance from each method.  In all cases, we applied the same data augmentation pipeline, in which we first randomly cropped the data and then randomly flipped the data horizontally.  All evaluations used a variant of stochastic gradient descent (SGD) for the optimization and minimized the cross-entropy loss.  We focused on linear probing, since that seemed to better highlight differences in performance between methods.

\textbf{SimCLR, BYOL, and MoCo v2:} For these methods, we set a single evaluation template because we found that they all more or less performed on par with their reported performance using this configuration.  We used SGD with a momentum of 0.9, Nesterov acceleration \citep{nesterov_dl,nesterov}, and no weight decay.  We used a batch size of 1024, and trained the linear classifier for 90 epochs after which we evaluated the performance on the validation set. Our learning rate followed a cosine decay schedule starting at 1.2 (0.3 scaled by the batch size) and decaying to 0.

\textbf{SwAV:} We found that SwAV performed significantly worse than reported when using this default evaluation protocol.  While the exact details of why are unclear, we found that using the hyperparameter configuration from \citet{swav} fixed most of the performance discrepancy.  We used SGD with a traditional momentum (i.e. no Nesterov acceleration) of 0.9 and a weight decay of $10^{-6}$.  We trained with a batch size of 256, a cosine decay learning rate schedule starting at 0.3 and decaying to 0, and we trained for 100 epochs.

\textbf{DINO:} We also found that DINO performed significantly worse than reported when using the default evaluation protocol.  In order to follow the published implementation, we used a concatenation of the cls token and the globally average pooled features from the last transformer block as the inputs to our linear classifier.  We used a batch size of 1024, SGD with Nesterov momentum and no weight decay for our optimizer, and a cosine learning rate schedule starting at 0.003 and decaying to 0.  We trained for 100 epochs.

\textbf{MoCo v3:}  We also found that MoCo v3 performed worse than reported when using the default evaluation protocol.  However, this appeared to be solved by increasing the learning rate significantly to 12.0 (3 scaled by the batch size),  and by using traditional momentum rather than Nesterov.

\subsection{Methodology for measuring performance improvements}\label{sec:computing_averages}
In the main text of the paper, we reported average improvements generated from enhancements to augmentations, algorithms, and architectures, and used those numbers to explain how much of recent advances in measured performance were caused by each. Here we detail the methods used to compute the average improvements, and also discuss alternative measurements. Our main objective is to assess how much of the performance gap between SimCLR and DINO is due to augmentations and how much is due to algorithms (which also requires us to assess how much is due to architectures). To assign relative importance to each, we measure the impact of augmentations, architectures, the momentum encoder, and the prediction network individually. We then identify how much of the gap between SimCLR and DINO is explained by these 4 measurable changes.  The remainder is assumed to be attributed to the pretraining task

\textbf{Augmentations:} In general, we measure the improvement caused by an augmentation to be the increase in the average top 1 accuracy produced from switching augmentations.  We then report the average increase computed over methods.  We observed an average improvement of 14\% between Crop and Crop + Color augmentations, an average improvement of 3.4\% between Crop + Color and SimCLR augmentations, an average improvement of 2.3\% between SimCLR and BYOL augmentations, and an average improvement of 2.5\% between BYOL and multi-crop augmentations.  The average improvement from Crop to multi-crop is 23.2\%, and the average improvement from the least performant augmentation to the best is 23.4\%.  The average improvement from SimCLR to multi-crop is 4.9\%, while the average improvement from SimCLR to the most performant augmentation is 5.2\%.  We therefore reported the improvement to the nearest percentage for the total improvement from SimCLR to multi-crop, since our uncertainty is on the order of 0.1\%.

\textbf{Algorithms:} Algorithms were slightly more tricky to measure because in some cases the algorithm was necessary to avoid representation space collapse. Therefore, if we were to take the average over our momentum encoder and predictor ablations, we would find that momentum encoders constitute an improvement of $10\pm28\%$, while prediction networks improve performance by $1\pm32\%$, which would make it impossible to draw any meaningful conclusions about their impact.  However, we also note that we are trying to assess the portion of the improvement from SimCLR to DINO that was driven by algorithmic advances.  Therefore we removed cases where either baseline performed worse than the SimCLR baseline, since those cannot be used to explain improvements in performance relative to SimCLR.  This results in computing the momentum encoder improvement from SimCLR, MoCo v2, DINO (with no predictor) and MoCo v3, over which we measure a 1.8\% average improvement.  This similarly results in computing the predictor improvement from SimCLR, MoCo v2, SwAV (with no momentum), DINO (with momentum), and MoCo v3, over which we measure an average improvement of 0.3.  We could also exclude clustering methods from this analysis, in which case we would measure an average improvement of 0.5, however since the top performing method is a clustering method, it seems reasonable to include them in the averages.

\textbf{Architectures:} Architectures were the most difficult comparison to make, mainly because there were relatively few cases where the only change between methods was the architecture.  The comparisons we used were between MoCo v3 and MoCo v2, and between SwAV and DINO, where we used our DINO experiment that did not use a momentum encoder.  For MoCo, we took the average improvement over augmentations as the measured improvement, and averaged that with the improvement we observed for DINO to equally weight the improvements from different techiques. We find the average improvement to be 2.3\%.  If we instead marginalize over augmentations, taking the best performing augmentation for each architecture and computing their difference, we find the architecture-induced improvement averaged across methods to be 2.2\%.  We therefore report 2.3\% as the improvement from architectures.  We also note that the average improvement seen for contrastive tasks (3.3\%) is larger than that seen for clustering tasks (1.2\%), thus weighting each possible comparison equally would suggest a 2.8\% improvement from improving the architecture. 
\section{Additional Experiments}
In this section, we report a few small additional experimental results.  Our focus with these experiments was to perform ablations on top of our baselines in order to ascertain the impact of certain hyperparameters.

\subsection{DINO enhancements}
DINO is the most performant of the methods we consider.  This is a convergence of many individual improvements to the modeling, including a larger model architecture than its predecessor, SwAV an additional couple of local views, and schedules for weight decay and teacher temperature. Relative to the concurrent work from \citet{mocov3} it also has minor differences.  Our goal was to assess the impact that each of these differences has on performance. In order to do this, we ran several experiments with DINO dropping various aspects of the configuration to use the setup that would mirror other works.  Those experiments included 1) Removing the weight decay schedule, 2) Removing the temperature schedule, 3) Removing stochastic depth, 4) Using Relu in the projection head, 5) Removing clipnorm from the optimizer and 6) Using a larger batch size commensurate with other methods.

\begin{table}[ht]
    \centering
    \begin{tabular}{l|c}
        \toprule
        Hyperparameters & Top-1 Accuracy  \\
        \midrule
        Baseline & 78.0\\
        Fixed weight decay (0.04) & 74.5\\
        Fixed weight decay (0.1) & 76.9 \\
        Fixed weight decay (0.4) & 76.5 \\
        Fixed Temperature (0.025) & 77.7 \\
        Fixed Temperature (0.05) & 77.9 \\
        No Stochastic Depth & 78.2 \\
        No Gradient Clipping & 77.5 \\
        ReLU Projector & 77.4\\
        Batch Size = 2048 & 0.1\\
        \bottomrule
    \end{tabular}
    \caption{\textbf{Ablations on Dino:} Impact on downstream task performance of DINO as a function of the embellishments used during pretraining.}
     \label{tab:dino_ablations}
\end{table}

Each of these experiments kept all other hyperparameters fixed using the configurations detailed in \cref{sec:hyperparameters}.  The specific change to the config, and the measured performance is reported in \cref{tab:dino_ablations}.  We also make a few observations from these experiments.  First, from \cref{tab:augmentations} we see that the natural variance between experimental runs for dino is 0.2, meaning that any performance degradation or enhancement needs to be significantly in excess of this amount in order to be considered meaningful.  With this in mind, we find that the temperature schedule does not appear to have a significant impact on performance, where fixing it at both the initial temperature and the final temperature only observe minor performance degradations easily explainable via random scatter.  We also find that removing stochastic depth is consistent with having no impact on performance.  Clipnorm also appears to have only a minor impact, lowering performance only by a 0.5\% relative to the baseline.  Using ReLU instead of GELU in the projector lowers performance by 0.6\%, slightly larger than clipnorm but still not particularly significant.  However, using a fixed weight decay does appear to have a statistically significant impact, lowering performance by 1.1\% in the optimal case, and by 3.5\% in the worst case.  Finally, we find that DINO was not able to train successfully with batch sizes larger than 2048, achieving representation space collapse and getting 0.1\% accuracy. Interestingly, this particular ablation was not included in the DINO paper (which only considered smaller batch sizes than the default), so it is difficult for us to tell if the collapse is intrinsic or just due to the particular combination of hyperparameters that we used.

\subsection{MoCo v3 temperature tuning}
MoCo v3 is the second most performant of the methods we consider.  One of our hypotheses in this work is that the pretraining task is not a significant driver of performance. We found in our experiments that while MoCo v3 performs better than was reported in \citet{mocov3}, it also appears to lag behind the observed performance from DINO, slightly undermining this conclusion. At the same time, we were also puzzled that MoCo v3 appeared to raise the loss temperature, since we had found in previous experiments that the higher temperatures performed worse for other methods like SimCLR.  We therefore ran an additional experiment lowering the temperature while keeping all other elements of the experiment configuration fixed.  We ran a single trial, and found that this run observed 78.2\% accuracy on ImageNet, consistent with our results from DINO, and suggesting that the gap between them was mainly a consequence of improper tuning for MoCo v3.

\end{document}